\documentclass{article}

\usepackage{microtype}
\usepackage{graphicx}
\usepackage{subfigure}
\usepackage{booktabs} % for professional tables

\usepackage{hyperref}

\usepackage[accepted]{icml2020}

\icmltitlerunning{Natural Language State Representation}

\usepackage{multirow}

\begin{document}

\twocolumn[
\icmltitle{An Overview of Natural Language State Representation \\
           for Reinforcement Learning}

% It is OKAY to include author information, even for blind
% submissions: the style file will automatically remove it for you
% unless you've provided the [accepted] option to the icml2020
% package.

% List of affiliations: The first argument should be a (short)
% identifier you will use later to specify author affiliations
% Academic affiliations should list Department, University, City, Region, Country
% Industry affiliations should list Company, City, Region, Country

% You can specify symbols, otherwise they are numbered in order.
% Ideally, you should not use this facility. Affiliations will be numbered
% in order of appearance and this is the preferred way.
\icmlsetsymbol{equal}{*}

\begin{icmlauthorlist}
\icmlauthor{Brielen Madureira}{up}
\icmlauthor{David Schlangen}{up}
%\icmlauthor{Cieua Vvvvv}{goo}
%\icmlauthor{Iaesut Saoeu}{ed}
%\icmlauthor{Fiuea Rrrr}{to}
%\icmlauthor{Tateu H.~Yasehe}{ed,to,goo}
%\icmlauthor{Aaoeu Iasoh}{goo}
%\icmlauthor{Buiui Eueu}{ed}
%\icmlauthor{Aeuia Zzzz}{ed}
%\icmlauthor{Bieea C.~Yyyy}{to,goo}
%\icmlauthor{Teoau Xxxx}{ed}
%\icmlauthor{Eee Pppp}{ed}
\end{icmlauthorlist}

\icmlaffiliation{up}{University of Potsdam, Potsdam, Germany}
%\icmlaffiliation{goo}{Googol ShallowMind, New London, Michigan, USA}
%\icmlaffiliation{ed}{School of Computation, University of Edenborrow, Edenborrow, United Kingdom}

\icmlcorrespondingauthor{Brielen Madureira}{madureiralasota@uni-potsdam.de}
%\icmlcorrespondingauthor{}{david.schlangen@uni-potsdam.de}

% You may provide any keywords that you
% find helpful for describing your paper; these are used to populate
% the "keywords" metadata in the PDF but will not be shown in the document
\icmlkeywords{NLP, Reinforcement Learning, natural language, state representation, overview}

\vskip 0.3in
]

% this must go after the closing bracket ] following \twocolumn[ ...

% This command actually creates the footnote in the first column
% listing the affiliations and the copyright notice.
% The command takes one argument, which is text to display at the start of the footnote.
% The \icmlEqualContribution command is standard text for equal contribution.
% Remove it (just {}) if you do not need this facility.

\printAffiliationsAndNotice{}  % leave blank if no need to mention equal contribution
%\printAffiliationsAndNotice{\icmlEqualContribution} % otherwise use the standard text.

\begin{abstract}
A suitable state representation	is a fundamental part of the learning process in Reinforcement Learning. In various tasks, the state can either be  described by natural language or be natural language itself. This survey outlines the strategies used in the literature to build natural language state representations. We appeal for more linguistically interpretable and grounded representations, careful justification of design decisions and evaluation of the effectiveness of different approaches.	
\end{abstract}

\section{Introduction}
\label{intro}

In any typical Reinforcement Learning (RL) scenario, there is an agent in an environment following a policy to take actions which make it transition through states and receive rewards~\cite{sutton1998reinforcement}. Each of these elements must be modeled according to the purpose of the task and their representation can influence the outcome of RL algorithms. Natural Language Understanding can be integrated into these components, both in \textit{language-conditional} RL (when natural language is inherent to the task) and in \textit{language-assisted} RL (when natural language is an additional tool), as discussed in~\citet{luketina2019survey}.

\begin{figure}[ht]
	\vskip -0.1in
	\begin{center}
		\centerline{\includegraphics[trim={0.5cm 22cm 3cm 0cm},clip,width=7.5cm]{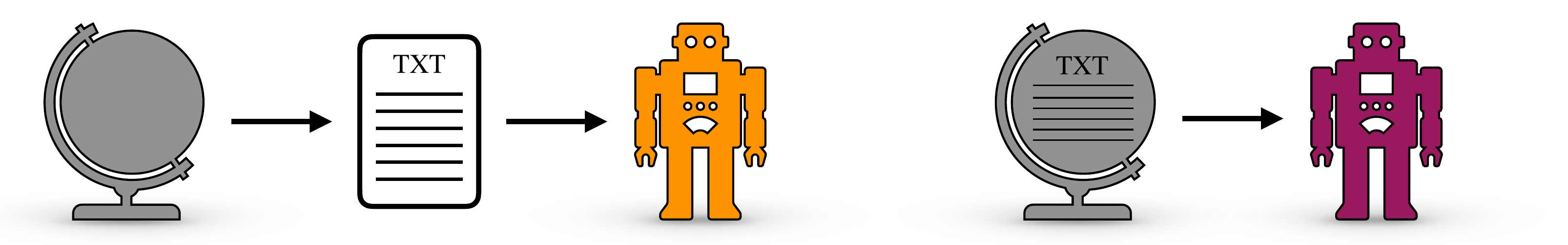}}
		\caption{Natural language can describe the agent's state (\textit{e.g.} text-based games) or be part of it (\textit{e.g.} dialogue or text summarization).}
		\label{diagram}
	\end{center}
	\vskip -0.4in
\end{figure}

The reward function is usually under the spotlight because the agent's goal is maximizing the expected long-term reward. But the state representation is no less vital. Based on it, the agent perceives the environment and comes to decisions on how to act. While in robotics the agent observes a spatial environment and in arcade games the state may be composed of a sequence of images, natural language is as well a common part of RL states, as illustrated in Figure~\ref{diagram}. 

The choice of state representation is a problem on its own. It directly affects the learning process~\cite{jones2010integrating}, so if applications neglect this, the agent may be prevented from accessing key information for decision-making.

There is ongoing research on the use of natural language to model the reward and the actions~\cite{he-etal-2016-deep,feng2018extracting,goyal2019using}, but we are not aware of any recent systematic survey about the various possibilities of using natural language to model the state representation and how to do that effectively. We aim at filling this gap by providing an overview of previous work in which the state is based on text, delineating the main approaches.

We dive into a wide range of NLP papers that apply RL methods and whose state representations have to capture linguistic features that influence decision-making. Findings about state representations in this area may potentially be extrapolated to other language-informed RL tasks. We thus hope this overview aids those seeking the objective of having RL agents capable of understanding natural language. Since we notice there is no consensus on how to design natural language state representations, we conclude with some concrete recommendations for future research.

\section{State Representation in RL}

The construction of suitable state representations for RL has been assessed by several works, for example in the field of robotics and autonomous cars, where modeling the state is particularly hard due to very high dimensional data coming from multiple sensors~\cite{jonschkowski2013learning,de2018integrating}.~\citet{lesort2018state}, for instance, presented a review on State Representation Learning algorithms and evaluation methods for control scenarios.

By definition, the state in a Markov Decision Process, which underlies the formalization of RL, needs to be Markovian; \textit{i.e.} all that needs to be known about the past must be contained in the current state.~\citet{jonschkowski2013learning} summarized other desirable properties of state representations proposed by many authors: It should provide good features for learning the value function; be compact but still allowing the original observations to be reconstructed; change slowly over time; be useful for predicting future observations and rewards given future actions; and ideally be shared by various similar tasks. Although their focus was robotics, these properties can be extended to textual states. 

Many other aspects of state representation have been investigated.\footnote{See \textit{e.g.}~\citet{mccallum1996hidden,kaelbling1998planning,finney2002thing,van2002relational,morales2004relational,roy2005finding,frommberger2008learning,mahmud2010constructing,maillard2011selecting,ortiz2018learning,franccois2019combined}.} In early works, states were handcrafted by the system designer using features. Although it remains a common practice, feature engineering has evident drawbacks: it needs an experienced designer, is tedious and time-consuming and does not generalize~\cite{bohmer2015autonomous}. The popularization of deep learning can to some extent spare us part of these shortcomings, as raw data can potentially be used as input to end-to-end models that build state representations implicitly.~\citet{de2018integrating}, for instance, analyzed methods for integrating state representation learning into deep RL. Still, choosing which information, data structure and model to use can be regarded as a modern form of feature engineering. Several options have been applied to textual data, as we describe in the next section.

\section{State Representation in NLP tasks}

When RL methods are adopted to solve NLP tasks, one common characteristic is that the state $S$ is represented as a function of texts $T$, \textit{i.e.} $S=f(T)$. As we will see, $f$ may take a variety of forms and additional inputs. We begin by reviewing the development of natural language state representation in dialogue, instruction following, text-based games and text summarization, NLP areas in which RL methods have been used more often. With the consolidation of deep RL methods, more studies tried to solve several other tasks, which we aggregate in the last part.\footnote{LSTM~\cite{hochreiter1997long} and GRU~\cite{cho2014learning} are references to models mentioned throughout.}

\textbf{Dialogue} is likely the NLP area with the most substantial number of papers that adopt RL, with research going on for more than two decades, \textit{e.g.}~\citet{levin1998using}. In dialogue, the state representation is usually a mapping from text to an abstract or compact format, such as a template, a belief state or an embedding vector, used by the agent to generate the next utterance. A Natural Language Understanding component between the text and the state representation is common, as illustrated, for example, in~\citet{lipton2018bbq}. 

Delineating the state space was at first mostly a manual task, because considering the entire dialogue used to be impractical~\cite{singh2000empirical}. Keeping the state space small was a common concern in order to avoid the curse of dimensionality~\cite{levin2000stochastic,singh2000empirical}. Therefore, system designers picked a set of variables or features based on their experience, usually resulting in application-dependent representations that did not generalize~\cite{walker2000application,levin2000stochastic,frampton2005reinforcement,english-heeman-2005-learning,mitchell-etal-2013-evaluating,khouzaimi-etal-2015-optimising}. The slots and values of the dialogue state were used to represent the environment state~\cite{papangelis-2012-comparative} or the Information States approach was incorporated~\cite{georgila2005learning,henderson2005hybrid,heeman2012using}. 
It was soon clear that modeling the state space is a fundamental aspect of RL for dialogue, as it has direct impact on the dynamics of the system, and attention was drawn to the fact that the research community had neither established best practices for modeling the state nor agreed upon domain-independent variables~\cite{paek2006reinforcement}. 

Some studies tried to examine which state representation was more effective for the main task~\cite{scheffler2002automatic}, constructing them with different feature combinations. There was also discussion on ensuring the Markovian property~\cite{singh2000reinforcement}, which means that the ``representation must encode everything that the system observed about everything that has happened in the dialogue so far"~\cite{walker2000application}. Ablation studies and metrics to evaluate the effect of each feature were proposed~\cite{frampton2005reinforcement,tetreault2008reinforcement,heeman2009representing,mitchell-etal-2013-evaluating}. 

Once neural network models started to be employed, new ways to represent states and to integrate representation learning into the agent's learning were enabled. States could be more easily represented by a vector and fed directly into a parametrized value or policy function. A common approach was building a belief state (a distribution over possible dialogue states, usually represented by a fixed set of slot-value pairs) when dialogue tasks are modeled as POMDPs~\cite{su2016continuously,fatemi2016policy,wen2017network,weisz2018sample}. In~\citet{wen2017network}, a sequence of free form text was mapped into a fixed set of slot-value pairs by an RNN.~\citet{dhingra2017towards} compared a handcrafted and a neural belief tracker that uses a GRU over turns. The belief state was sometimes combined with other representations.~\citet{wen2017latent} modeled the dialogue state vector as a concatenation of user input (encoded by a BiLSTM), a belief vector (probability distributions over domain specific slot-value pairs, extracted by RNN-CNN belief trackers) and the degree of matching in a knowledge base.

Note that, even when the designer does not have to build the representation manually, the selection of slots and value ranges, the choice of input features and the NLU representations still require human intervention. Features can still be used as input to deep learning models.~\citet{williams2016end} represented the dialogue history with features (such as input and output entities) as input to an LSTM, which inferred the state representation and mapped the input directly to actions. The NLU component played a key role in the state representation \textit{e.g.} in~\citet{manuvinakurike-etal-2017-using}.

\citet{cuayahuitl2016deep} pointed out that it is typically unclear what features to incorporate in a multi-domain dialogue and proposed applying deep RL directly to texts, so that the agent learns feature representation and the policy together, bypassing NLU components. Similarly,~\citet{li-etal-2016-deep} used the two previous dialogue turns, transformed into a continuous vector representation by an LSTM encoder, to build representations directly from raw text.

A common strategy has been to use the hidden vector of RNNs to represent and update the environment state~\cite{zhao2016towards,williams2017hybrid,liu2017iterative}. Multimodal state representation has also become frequent, combining encoded text with images~\cite{Das-2017-ICCV}, knowledge base queries~\cite{li2017end,liu2017iterative} or an embedding of a knowledge graph~\cite{yang2020reinforcement}. 

In \textbf{Instruction Following}, deciding the next action means directly interpreting a textual instruction. In this task, states were first designed as tuples of world and linguistic features: words and documents~\cite{branavan-etal-2009-reinforcement,branavan2010reading} or cardinal directions and utterances~\cite{vogel2010learning}. We then observed the use of sentence embeddings (usually by an~LSTM or~GRU) combined with other sources such as image embeddings~\cite{misra2017mapping,hermann2017grounded,kaplan2017beating,fu2018from} or instruction memory~\cite{oh2017}. A state processing module was introduced in~\citet{chaplot2018gated} to create a joint representation of the image and the instruction.~\citet{janner2018representation} converted the instruction text into a real-valued vector and used it both to build a global map-level representation and as a kernel in a convolution operation to obtain a local representation of the state.

In the related vision-language navigation task, the state was modeled by combining visually grounded textual context and textually grounded visual context~\cite{wang2019reinforced,wang2020visionlang} or by using an attention mechanism over the representations of the instruction~\cite{anderson2018vision}.

\textbf{Text-Based Games} pose a related, but more general challenge, as here the text describing the current game state must be integrated into the agent state and the best action must be inferred. The work of~\citet{narasimhan-etal-2015-language} was a reference for modeling the state representation in text-based games, which is jointly learned together with action policies in their setting. They used an LSTM over textual data with a mean pooling layer on top, in an attempt to capture the semantics of the game states. Subsequent works followed the same approach~\cite{ansari2018language,yuan2018counting,jain2019algorithmic} or used variants of continuous vectors built by neural networks~\cite{he-etal-2016-deep, foerster2016}.

Other elaborate ideas appeared subsequently.~\citet{narasimhan2018grounding} used a factorized state representation, concatenating an object embedding with its textual specification embedding (LSTM or bag-of-words). ~\citet{ammanabrolu-riedl-2019-playing} combined the embedded text (by a sliding BiLSTM) with an embedded knowledge graph built by the agent throughout the game. Their concatenation served as input to an MLP that outputted state representations.~\citet{murugesan2020enhancing} used embeddings of a local belief graph and a global common sense graph of entities.~\citet{zhong2019rtfm} proposed building representations that capture interactions between the goal, a document describing environment dynamics, and environment observations.

\textbf{Text Summarization} is another flourishing area for RL. In this task, an agent processes a text and either pick key sentences to compose a summary or use Natural Language Generation to output its own words. Initial works applying RL used tuples of features to represent the state, composed of the summary at each time step, a history of actions and a binary variable indicating the terminal state~\cite{ryang2012framework,rioux2014fear,henss2015reinforcement}. 

RL was really consolidated for text summarization after deep learning methods became available. In most cases, the state representation was, as in other tasks, the hidden state of an RNN generating the summary~\cite{ling2017coarse,pasunuru2018multi}, also together with an encoding of a candidate sentence~\cite{lee2017automatic}.~\citet{paulus2018a}, a reference for subsequent works~\cite{kryscinski2018improving}, used context vectors with intra-temporal attention over the hidden states of the encoder (which process the original text) concatenated to the hidden state of the decoder (which generates the summary).

Hierarchical approaches are common.~\citet{wu2018learning} built a document encoding with a CNN operating at word level and a BiGRU at sentence level, whereas~\citet{narayan2018ranking} and~\citet{chen2018fast} used a CNN at sentence level and an LSTM or BiLSTM at document level to capture global information.  Likewise,~\citet{yao2018deep} employed an RNN or CNN sentence encoder, together with a representation of the current summary and the document content.

\textbf{Other NLP Tasks} comprise each a currently smaller number of studies using RL, so we group them here into main general choices of state design. Some tasks used ordered words~\citep[paraphrase generation]{li-etal-2018-paraphrase}, predicted words~\citep[simultaneous machine translation]{grissom-ii-etal-2014-dont}, a vocabulary set and a taxonomy~\citep[taxonomy induction]{mao-etal-2018-end} or a bag of sentences~\citep[relation extraction]{zeng2018large}. Coreference resolution adopted partially formed coreference chains~\citep{stoyanov-eisner-2012-easy} or word embeddings and features~\citep{clark-manning-2016-deep}. For syntactic or semantic parsing, the parser configuration was used as set of discrete variables describing the state of a parse structure~\cite{zhang-chan-2009-dependency,jiang2012parsing,le-fokkens-2017-tackling}, a concatenation of their representation built by an LSTM~\cite{naseem-etal-2019-rewarding} or a query for a knowledge base~\cite{liang-etal-2017-neural}. 

A usual strategy to represent the state was handcrafting vectors composed of selected variables or metrics, for example, similarity scores and number of words~\citep[text-based clinical diagnosis]{ling-etal-2017-learning}, confidence scores, td-idf and one-hot encoding of matches~\citep[information extraction]{narasimhan-etal-2016-improving,taniguchi-etal-2018-joint}, situational and linguistic information~\citep[NLG]{dethlefs-cuayahuitl-2011-hierarchical}, entitites and relations~\citep[question answering]{godin-etal-2019-learning}, probability distribution over a set of objects~\citep[question selection]{hu-etal-2018-playing} or values generated by a parser~\citep[math word problem]{wang2018mathdqn}. 

There has been a myriad of creative efforts to build representations using neural networks, especially RNNs, as one can seamlessly regard the RNN as an agent and its hidden state as the environment state. The internal state (hidden vector and/or cell vector in LSTMs) was set as the environment state in language modeling~\cite{ranzato2015sequence}, image captioning~\cite{rennie-2017-imagecap}, math word problem~\cite{huang-etal-2018-neural} and NLG~\cite{yasui-etal-2019-using}, sometimes with attention over hidden states of a sequence in grammatical error correction~\cite{sakaguchi-etal-2017-grammatical} or a concatenation of hidden states further encoded by another neural network in sentence representation~\cite{yogatama2017learning}. A Transformer's hidden states~\cite{vaswani2017attention} were also used in machine translation~\cite{wu-etal-2018-study}. The encoding provided by the output layer was used for semantic parsing~\cite{guu-etal-2017-language} and also with location-based attention for text anonymization~\cite{mosallanezhad-etal-2019-deep}. Hierarchical network representations combining word level and sentence level encodings were proposed, as in text classification~\cite{zhang2018learning}. Both the hidden state of the encoder and the decoder comprised the state in sentence simplification ~\cite{zhang2017sentence}. CNNs representations combined with predictive marginals were used in active learning for NER~\cite{fang-etal-2017-learning}.

\section{Concluding Discussion}

\begin{figure}[h]
	%\vskip -0.075in
	\begin{center}
		\centerline{\includegraphics[trim={0cm 18cm 15.5cm 0cm},clip,width=7.5cm]{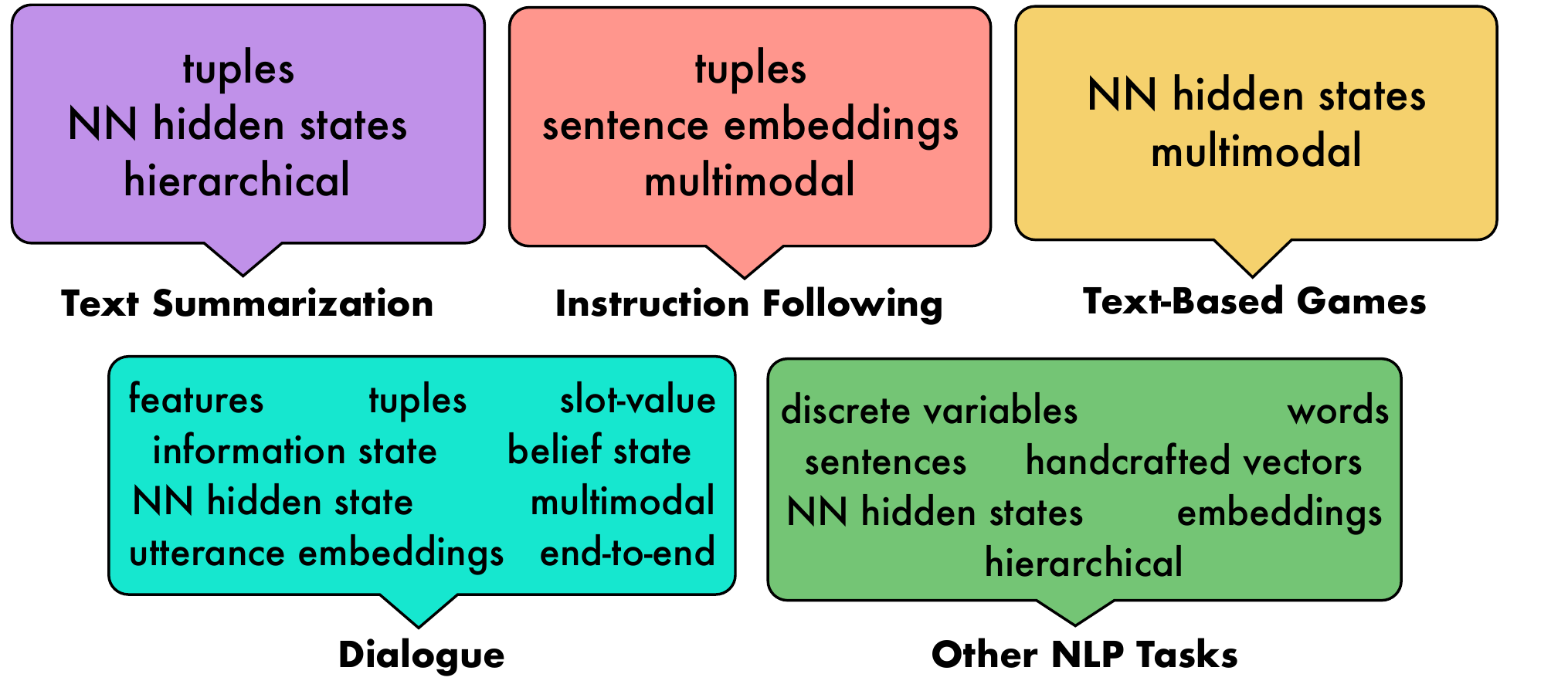}}
		\caption{Key concepts of state representation in each area.}
		\label{overview}
	\end{center}
	\vskip -0.4in
\end{figure}

The prolific literature combining NLP and RL reveals a noticeable synergy between them. While RL methods have extended frontiers of NLP research, natural language can be a valuable source of information for RL agents, in particular to model the state signal. An abundance of tailored natural language state representations has been explored and there is a current trend to opt for end-to-end models, encoding or decoding text with the aid of RNNs and for multimodal or hierarchical representations, as shown in Figure~\ref{overview}. 

More insight is needed on how natural language state representations should be put together. Unfortunately, it is not uncommon to find papers with unclear or missing definitions of their state signal or without much justification of their choices, which hinders comparison and reproducibility. We thus encourage researchers to carefully describe the state representation, provide justification for state design modeling and to also report attempts that had negative impact on the agent's learning process. 

Studies to validate the effectiveness of NL state representation used to be common when engineering features was a trend. Deep learning methods may give us the illusion that almost no human intervention or manual feature selection takes place, but they still play a relevant role in state signal design in terms of choice of architecture, model and input data. Adjusting evaluation and ablation methods to new approaches is thus necessary.~\citet{goyal2019}, for instance, experiment with three different natural language instructions representations, to test the effect of language-based reward. Similar analyses can also be done with language-based states, as in~\citet{narasimhan2018grounding}.

There is cutting-edge research being conducted about the interpretability of neural NLP models and their linguistic representational power~\cite{belinkov-glass-2019-analysis}. Natural language state representation would thus profit from studies about interpretability and also from diagnostic research~\cite{hupkes2018visualisation} on their abilities of distilling, composing and retaining semantic information throughout the agent's steps, usually expressed by recurrence in the neural networks.

In addition, grounding the meaning of texts to the dynamics of the environment, as discussed in~\citet{narasimhan2018grounding}, is a promising area of future research. The authors delineate two needs that must be met in state design with language grounding: the representation should fuse different modalities and capture the compositional nature of language in order to map semantics to the agent's world. 

Finally,~\citet{narasimhan-etal-2015-language,narasimhan2018grounding} discuss the benefits of representations that are effective across different games and that enable policy transfer. Therefore, another desirable property of natural language state representation is cross-domain validity, so that their encoding of world knowledge can be exploited in various RL scenarios.

\section*{Acknowledgments}
We thank the two anonymous reviewers for their feedback and suggestions.

%\clearpage
\bibliography{example_paper}
\bibliographystyle{icml2020}

\end{document}